\documentclass[letterpaper, 10 pt, conference]{ieeeconf}  

\IEEEoverridecommandlockouts                              

\usepackage{array}
\usepackage{amsmath,amsfonts}
\usepackage[linesnumbered,ruled,vlined]{algorithm2e}
\usepackage{algorithmic}
\usepackage{graphicx}
\usepackage{textcomp}
\usepackage{xcolor}
\usepackage{caption} 
\usepackage{subfigure}
\usepackage{amsmath}
\usepackage[normalem]{ulem}
\usepackage{blindtext}
\usepackage{multirow}
\usepackage{pgfplots}
\usepackage{comment}
\usepackage{booktabs}
\usepackage[T1]{fontenc}
\usepackage{tikz}
\usetikzlibrary{shapes.geometric, arrows}
\usepackage[normalem]{ulem}

\title{\LARGE \bf

Active  Shadowing (ASD): Manipulating Perception of \\Robotic Behaviors via Implicit Virtual Communication
}
 
 \author{Andrew Boateng, Prakhar Bhartiya, Taha Shaheen and Yu Zhang
 \thanks{A. Boateng, P Bhartiya, T. Shaheen and Y. Zhang are with the 
 School of Computing and Augmented Intelligence, Ira A. Fulton Schools of Engineering, Arizona State University, Tempe, AZ 85281, USA. Email: {\tt\small $\{$aoboaten, pbhartiy, tashahee, yu.zhang.442$\}$@asu.edu}.}}

\begin{document}
\maketitle
\thispagestyle{empty}
\pagestyle{empty}

\begin{abstract}

Explicit communication is often valued for its {\it directness} in presenting information but requires attention during exchange, resulting in cognitive interruptions. On the other hand, implicit communication contributes to tacit and smooth interaction, making it more suitable for teaming, but requires inference for interpretation. 
This paper studies a novel type of {\it implicit visual communication} (IVC) using shadows via visual projection with augmented reality, referred to as active shadowing (ASD). 
Prior IVC methods, such as legible motion, are often used to influence the perception of robot behavior to make it more understandable. 
They often require changing the physical robot behavior, resulting in suboptimality. 
In our work, we investigate how ASD can be used to achieve similar effects {\it without} losing optimality.   
Our evaluations with user studies demonstrates that ASD can effectively creates ``illusions'' that maintain optimal physical behavior without compromising its understandability. We also show that ASD can be more informative than other explicit communication methods, and examine the conditions under which ASD becomes less effective.

\end{abstract}

\section{INTRODUCTION}


Traditionally, explicit communication (EC) refers to verbal communication, whereas implicit communication (IC) involves using non-verbal cues. In the context of HRI, Che et al.~\cite{Implicit_explicit} expanded the definition of EC to include any form of communication—verbal or non-verbal—that conveys 
information via established channels and symbols so that the communicative intent is clear. 
Information communicated via EC is often designed to incur 
 minimal ambiguity, requiring little to no effort to interpret. Alternatively, IC is defined as all other communication
with intent that requires inference based on the current context. 
In such cases, the information communicated can be ambiguous, and that often implies more  effort in its interpretation.
Under such an expanded definition, non-verbal cues such as colors, lights, and sounds can be explicit communication (e.g., traffic lights).


While EC is valued for its directness in supporting information, it also introduces cognitive interruptions during transmission and reception since both the transmitter and receiver must divert their attention to the communication, which can negatively impact task fluency~\cite{Implicit_explicit}. In contrast, IC contributes to tacit and smooth interaction, which makes it more preferred when fluency is desired, such as in teaming and other applications where information communication must remain non-intrusive~\cite{fluent_interaction}.
In this paper, we study a novel type of implicit visual communication (IVC). Prior IVC methods, such as legible motion~\cite{GenLegible}, are often used to influence the perception of robot behavior to make it more understandable. However, they often require changes to the robot behavior, resulting in suboptimality. 
In our work, we propose the use of active shadows (ASD) to remove such a restriction. 
Furthermore, we hypothesize that ASD can be more informative than explicit communication methods especially in robotic applications characterized by continuous and sometimes peculiar motions. 

\begin{figure}
\centering
{\includegraphics[width=0.65\columnwidth, height=0.45\columnwidth]{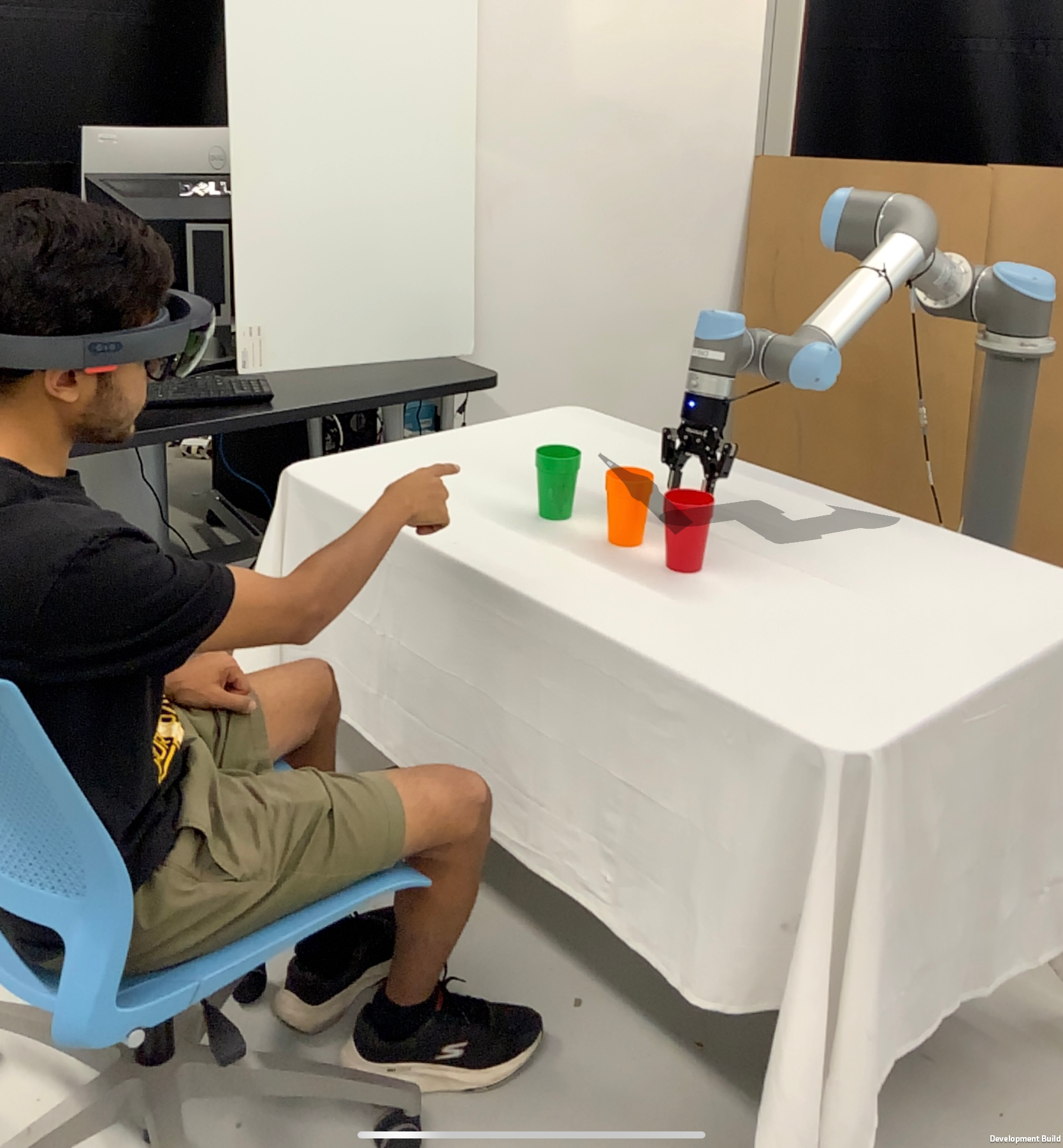}\label{fig:Shad_safety_b}}
\caption{Scenario that motivates the need for constant foreshadowing of the robot's behavior. In ASD, this is achieved by using virtual shadows to continuously project the robot's future state to assist in maintaining safer interaction.}
\label{fig:motivating}
\vskip-15pt
\end{figure}

Anyone who has operated a robot manipulator before would agree that it can generate unexpected behaviors. This is due to the different degrees-of-freedom and structure of the robotic manipulator, compared to human arms.
Consider the scenario shown in Fig. \ref{fig:motivating} where a human and a robot are tasked with cooperatively picking three evenly spaced cups from a shared workspace on a table. To ensure safety, the human must always keep the maximum distance from all parts of the robots, which can be achieved if the human always selects the cup that is farthest from the robot manipulator.  Assume that the robot goes for the cup in the middle. 
Even if the human is aware of the robot's goal, 
it would be difficult to safely choose a cup because the orientation of the manipulator during the pickup (i.e., the robot may {\it randomly} pitch left or right for the pickup, resulting in a close proximity from the butt end of the gripper to one of the cups on the side) determines which of the cups is safe for the human to pick up. 

Such a problem cannot be readily addressed by the existing IVC or explicit communication methods since such peculiarities can be difficult to foresee and make legible. 
Instead, a general solution would be a method that {\it constantly} foreshadows the robot's behavior without introducing  much cognitive interruptions, which is achieved in ASD.

In this paper, we propose Active Shadowing (ASD) to address the limitations of existing IVC methods by generating virtual shadows via augmented reality.
The intuition is that since shadows are often perceived as an integrative part of the object~\cite{illusion, IMURA2006652}, 
we can use shadows to manipulate the perception of the robot to introduce a discrepancy between the robot's physical behavior and its perception. 
Beneficially, it will allow us to maintain the optimal robot behavior without affecting its understandability,
leading to more efficient human-robot interaction. 
Furthermore, since the shadow represents an IC method and always accompanies the robot, its introduction would not require much attention.
It thus allows us to constantly foreshadow the robot's behavior and its various peculiarities, contributing to safer human-robot interaction. 
However, maintaining the association between the robot and its virtual shadow (as with its natural shadow) is challenging but crucial for effective manipulation. Such a requirement implies that the shadow must be seemingly ``natural'' so that its state can be attributed to the robot's state as desired. 
Our main contribution thus includes 1) introducing active shadowing as a novel IVC method with its unique characteriation: it bridges a gap between the existing IVC and EC methods, and 2) validating its effectiveness in manipulating perception and informativenness in controlled settings.
Note that although this work focuses on visual communication, the concept could extend to other non-verbal modalities such as sound and haptics.

\section{Related work}

\subsection{Communicating Robot Intent}
Effective HRI depends on clear and timely communication of the robot's intent, which influences human reactions and team collaboration~\cite{Comm_Intention_AR}. Improved communication enhances technology acceptance and situational awareness~\cite{mind, Situation_awareness}. Various methods exist for conveying robot intent, some of which are discussed below:

\subsubsection{Explicit Communication}
 Researchers have employed various communication modalities such as 3-D hand gestures~\cite{HandGesture}, facial expressions~\cite{BehavioralImplicit, Richardson2012EvaluatingHI}, body language~\cite{forceDetection}, acoustics or auditory~\cite{acoustical,distributed}, haptic~\cite{combine}, among others. 
Visual projections~\cite{heni, mind, light, light1} that are most relevant to our work have also been extensively studied. Technologies such as spatial projections and augmented reality (AR) enhance intent communication through visual overlays~\cite{intent_through_MR, spatial}. For example, to communicate robot intent, Chadalavada et al.~\cite{mind} proposed projecting the inner state information of a robot using a projector.
However, these methods typically belong to explicit communication that requires substantial attention, resulting in cognitive interruptions.


\subsubsection{Implicit Communication via Behavior Change}
A relevant line of work explores implicit communication (IC) through modifying robot behavior to convey intent or plans (hence also belonging to IVC). For instance, methods like ~\cite{GenLegible, intentExpressive, legPred, exp} alter the robot’s path to improve goal or plan inference. While effective in collaborative settings, these approaches often reduce task efficiency due to changes in the robot’s physical behavior. Such IC has also been  explored in adversarial settings for deception, as seen in emerging research on perception manipulation and deceptive behaviors~\cite{lie}. Techniques like goal obfuscation~\cite{Obfuscation} obscure objectives to protect information. 
However, as noted in~\cite{obfuscationAndLegibility}, changing behaviors can deceive both adversaries and teammates. 
One common problem with these prior IVC methods is that they introduce suboptimality since they directly change the behavior. 
Furthermore, it is sometimes difficult to apply them since it is not always clear when they are required, as discussed in the scenario in Fig. \ref{fig:motivating}.

\subsection{Influence of Shadows in Perception}
Shadows provide rich information about depth and motion of the associated objects~\cite{InduceDepth}. 
Research shows that an object's shadow can significantly influence the object's perceived trajectory ~\cite{illusion, IMURA2006652}, likely due to our expectation that shadows naturally move with their associated objects ~\cite{InduceDepth}. 
Shadow also enhances depth perception. For example, a cast shadow of a ball moving diagonally was found to give the impression of the ball receding in motion~\cite{Kersten}. Compared to other cues like light and sound, shadow conveys more information~\cite{objVsShadow}. Virtual shadows of robots have been shown to make them appear more natural and humane~\cite{Boateng, BoatengHRI}. 
These findings indicate that precise manipulation of an object’s shadow can effectively influence its perception, which motivates this work. Despite these early findings, the potential of using virtual shadows to create useful capabilities and applications in HRI remains under-explored and warrants investigation.

\section{Approach Overview} \label{approach}

\begin{figure}
\centering
\subfigure{\includegraphics[scale=0.1]{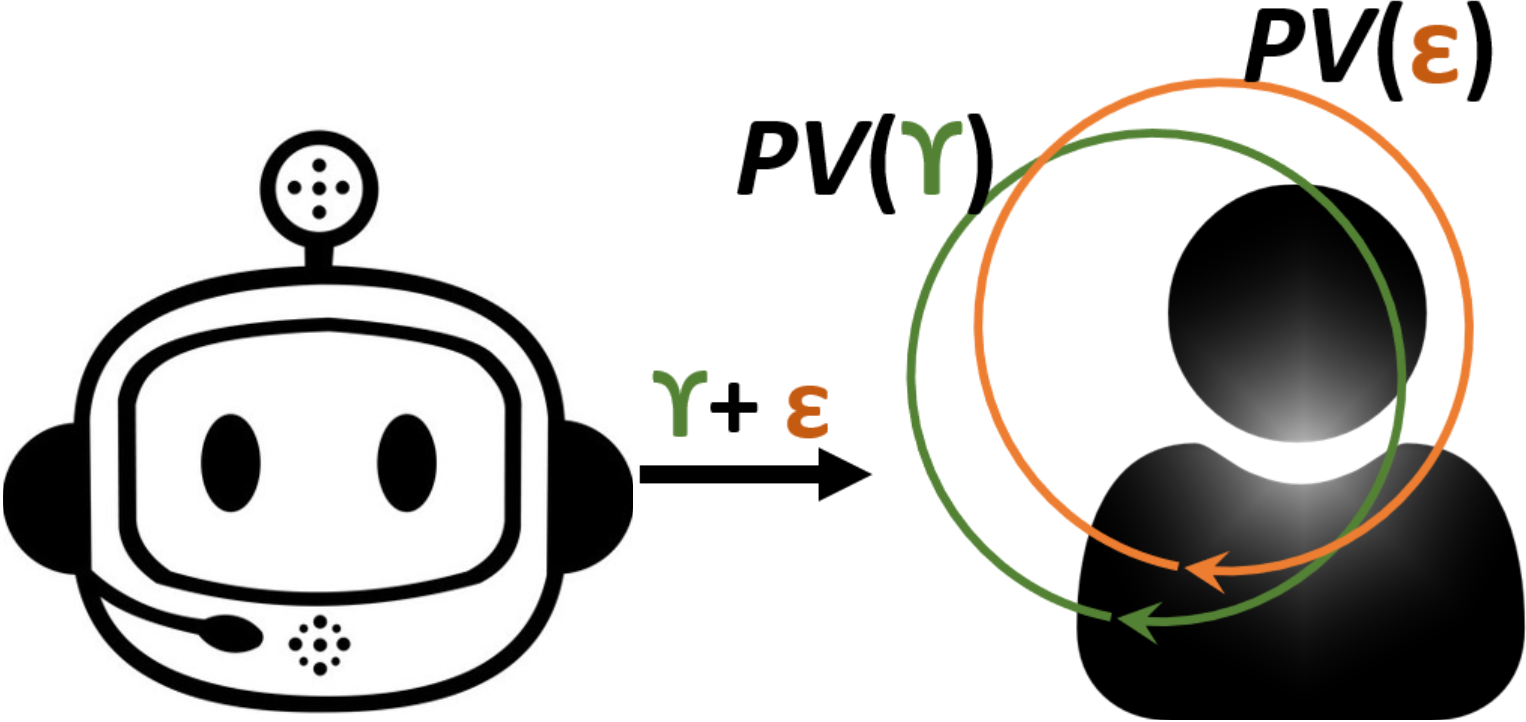}\label{illust_1}}
\subfigure{\includegraphics[scale=0.1]{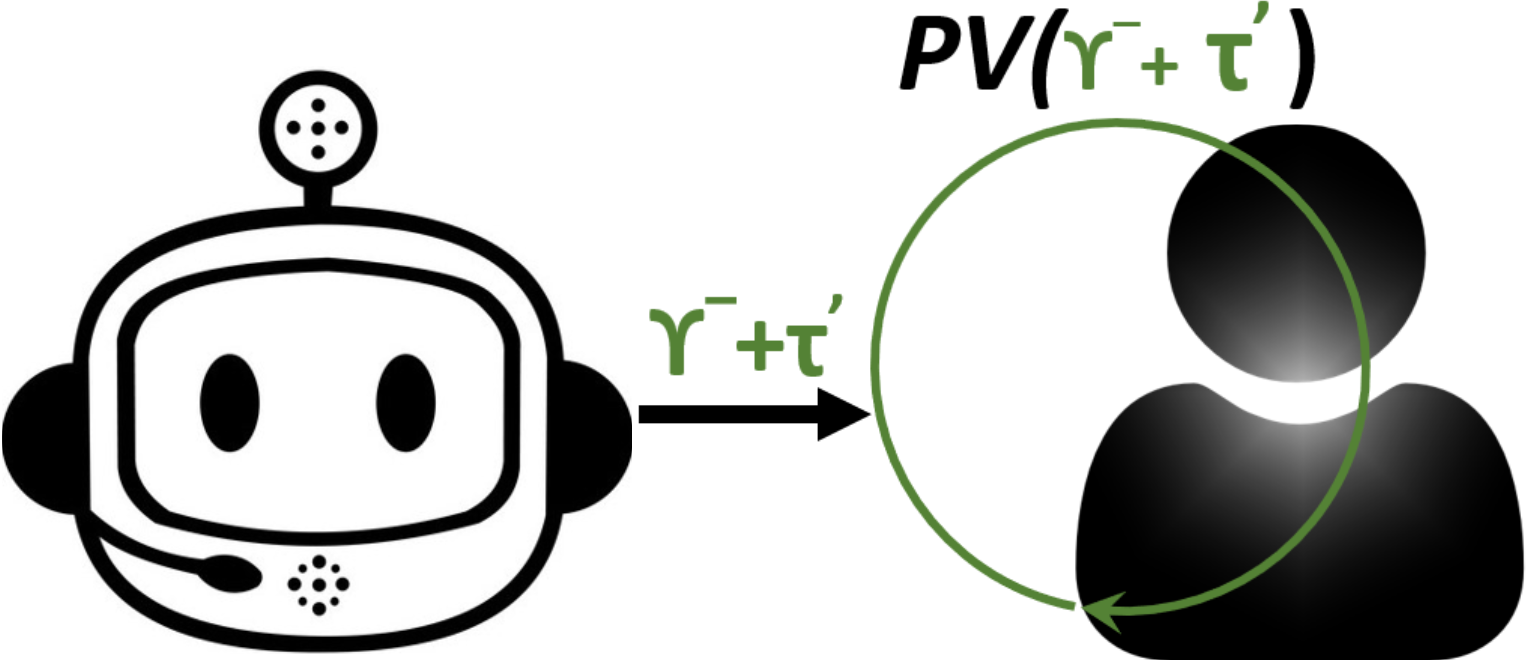}}\label{illust_2}
\vskip-5pt
\caption{
Illustration of the differences between ASD (right) and EC (left) and other IVC methods (directly changing $\gamma^-$).
}
\vskip-10pt
\label{fig:illustrative}
\end{figure}

The key characterization of active shadowing (ASD) that makes it unique are: 1) compared with prior IVC methods, the ability to affect perception without altering the physical robot behavior ($\gamma^-$ in Fig. \ref{fig:illustrative}), 
and 2) compared with explicit communication (EC) methods, 
it draws substantially less attention and is more suitable for continuous projection.
These differences are illustrated graphically in Fig. \ref{fig:illustrative}. 
With explicit communication, 
the human forms a perception of a robot ($\Sigma^+$) based on the robot's behavior (\(\gamma\)) and augmented information ($\epsilon$) that is explicitly communicated. This process can be represented as $\Sigma^+ = PV(\gamma) + PV(\epsilon)$.
In contrast, ASD decomposes $\gamma$ into two but integrative parts so that $\gamma$ = $\gamma^- + \tau$, 
where $\gamma^-$ denotes the physical behavior and $\tau$ is its naturally accompanying information, such as the robot's shadow. 
By modifying $\tau$ to $\tau'$, we can induce a desired perception $\Sigma^* = PV(\gamma^- + \tau')$ without altering $\gamma^-$,
resulting in an IC method since the communicative intent must be inferred. 
Prior IVC methods, such as legible motion, directly change $\gamma^-$ (which would also change $\tau$ naturally if present).

\begin{figure}[ht!]
\centerline{\includegraphics[scale = 0.12]{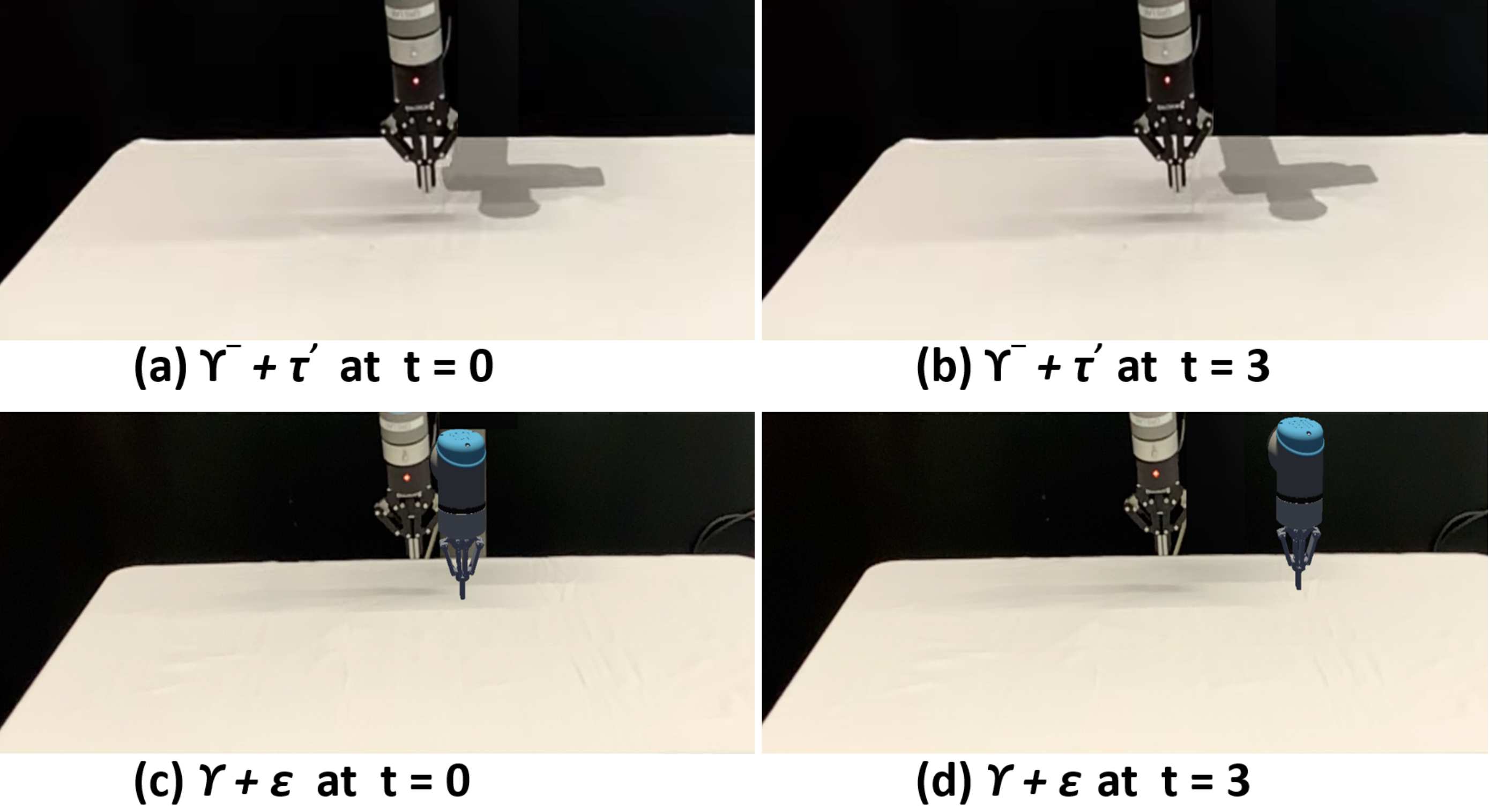}}
 \caption{
 \textbf{(a-b)} Manipulating the perception of behavior via ASD.
\textbf{(c-d)} Manipulating the perception of behavior using EC via projecting the hologram of the robot at a different position. In both cases, the robot did not move but the IC method provides a stronger illusion of motion.}
 \label{desired_trajectory}
 \vskip-10pt
\end{figure}

A requirement to create the discrepancy as illustrated in Fig. \ref{fig:illustrative} is the ability to modify the naturally accompanying information ($\tau$) of the original behavior ($\gamma$). There are two key considerations here: 1) ideally, we would like to make no changes to the physical behavior (e.g., physical trajectory) of the robot to maximize task performance, and 2) we must maintain the association between $\tau'$ and $\gamma^-$ (as with $\tau$ and $\gamma^-$) such that they can be interpreted integratively as with the original behavior $\gamma$. 
A specific case of $\tau$ we consider in ASD is  shadow 
since the original shadow $\tau$ can be effectively ``hidden'' by adding sufficient ambient light and
a visual shadow $\tau'$ can be projected in its place to restore such an association with the physical robot (see Fig. \ref{desired_trajectory}). 
Building on this intuition, we propose Active Shadowing (ASD) to generate such virtual shadows via augmented reality.

We leverage a prior framework~\cite{Boateng} to generate realistic virtual shadows to focus on the ``active'' part.   
The contribution of this work includes: 1) introducing ASD as a unique IVC method to manipulate perception, 
2) validating ASD's effectiveness through human subject studies using a physical robotic platform. Compared to the baselines, we demonstrate that ASD can maintain the optimal robot behavior while preserving its understandability. Additionally, our results show that ASD requires a comparable mental workload to the best performing baselines, thanks in part to our familiarity with interpreting shadows,
3) showing that ASD can be more informative than other explicit communication methods and analyzing the conditions under which such a visual manipulation becomes less effective. 

\section{Active Shadowing}
We make the following assumptions in this paper:
\begin{enumerate}
    \item Sufficient ambient light can be provided to reduce the effect of natural shadows.
    \item  The human is constantly monitoring the workspace (but potentially working on other tasks) that is augmented-reality enriched (e.g., human wearing an MS Hololens or operating a Collaborative Robot Workstation).
\end{enumerate}
\subsection{Preliminaries}
\subsubsection{Legibility}

Based on the psychological principle of ``action to goal'', Dragan et al.~\cite{GenLegible} mathematically generate a legible trajectory $\gamma^L$ that maximizes a trade-off between a legibility score $\xi$ and the trajectory cost of going to a goal $G$, where $\xi$ is computed as:
\begin{equation} 
    \xi = \frac{\int_0^T P(G|\gamma_{0:t})f(t)dt} {\int_0^T f(t)dt}
\,\,\,\,\,\,\,\,s.t. \,\,\,\, \gamma_T = G
    \label{eq:Anca}
\end{equation}
where $\gamma_{0:t}$ denotes a candidate trajectory from time $0$ to $t$,  
$P(G| \gamma_{0:t})$ is the probability of the robot moving towards the $G$ at time $t$, based on the  trajectory $\gamma_{0:t}$ observed so far.  
Trajectories are represented as vectors of waypoints and $\gamma_{t}$ denotes the waypoint at time $t$. 
$f(t)$ is a function that assigns greater weight to the early parts of the trajectory. 

\subsubsection{Realistic Virtual Shadows}
A prior work~\cite{Boateng} has developed a method to
project a virtual shadow of a robot along a trajectory ($\tau'$) via augmented reality,
which can deviate from the natural shadow's trajectory ($\tau$) along a given robot's trajectory ($\gamma^-$) 
by adjusting the pitch and yaw of the virtual light source (assuming directional light).
To ensure smooth and realistic shadow movements, a first-order discrete-time dynamic control model has been proposed there. In this paper, 
we use their approach to project our virtual shadow 
after determining its trajectory.
For simplicity, only the end point is considered in these trajectories (i.e., gripper tip). 

\begin{figure}[ht!]
\centerline{\includegraphics[width=0.45\columnwidth, height=0.35\columnwidth]{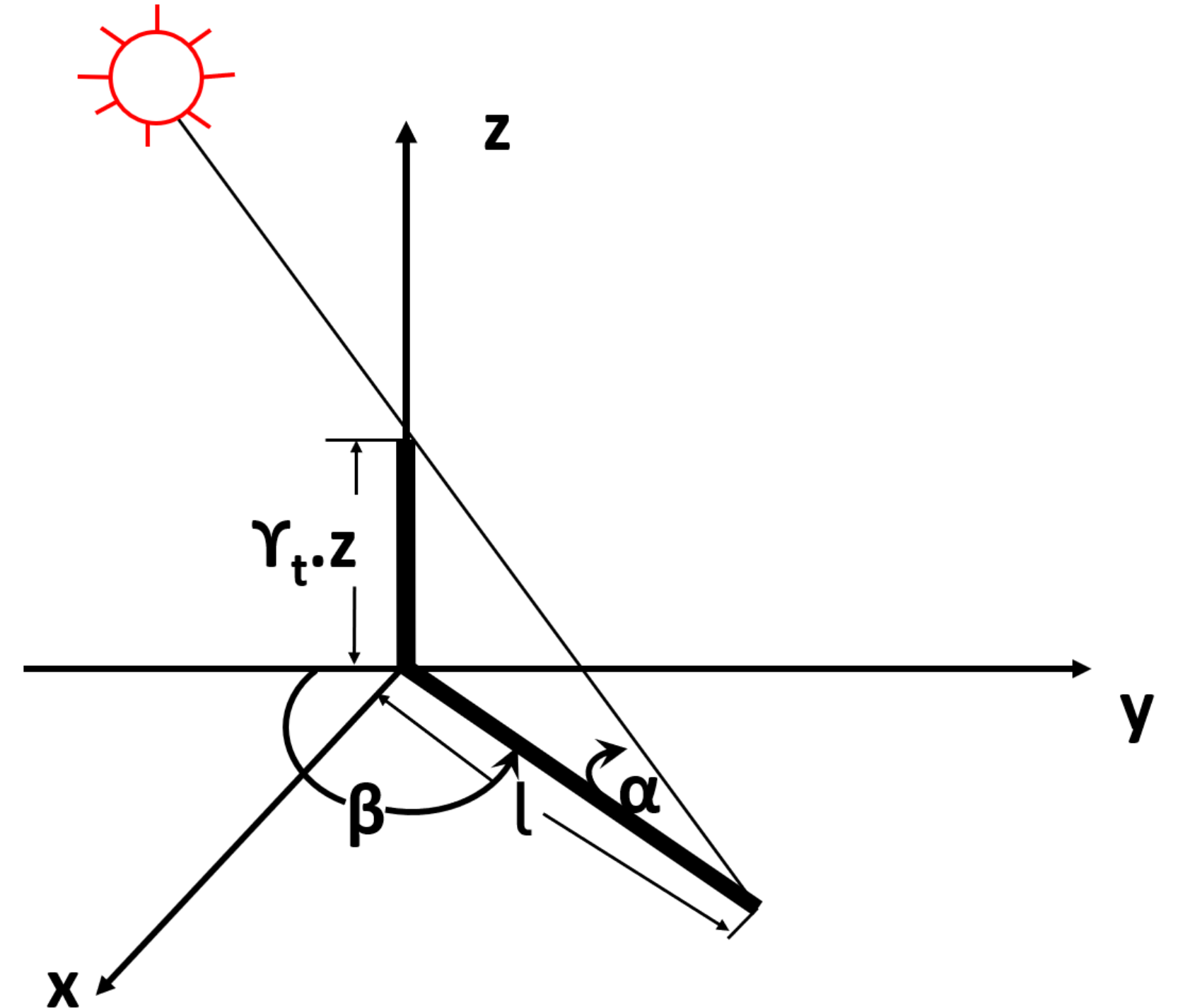}}
 \caption{Geometric representation of elevation/pitch ($\alpha$) and azimuth/yaw ($\beta$) of light source and their relation to shadow.}
 \label{sun}
\end{figure}
Given a robot trajectory \(\gamma\), the transformation function \(S(\gamma_t)\) converts each point along the robot's trajectory to its corresponding shadow point \(\tau_t\): \(\tau_t = S(\gamma_t)\). This transformation is defined by Eq.~\eqref{eq:manipulate} based on Fig. \ref{sun}:

\begin{equation}
    \begin{split}   
         \tau_t.x = - l \cdot \sin(\beta) + \gamma_t.x,   \tau_t.y = -l \cdot \cos(\beta) + \gamma_t.y
        \end{split}
    \label{eq:manipulate}
\end{equation} 
where $l = \frac{\gamma_t.z}{\tan(\alpha)}$, $(\gamma_t.x, \gamma_t.y, \gamma_t.z)$ are coordinates of the robot's gripper at time \(t\). Note that $\tau_t$ is 2D.

\subsection{Behavior Generation in ASD}

Fig.~\ref{fig:asd_model_flowchart}  summarizes the ASD process for generating virtual shadows. 
Given a cost-optimal robot trajectory $\gamma$ for achieving a given task and a desired perception of the robot's behavior \(\Sigma^*\), 
we can compute another trajectory $\gamma^*$ that deviates from $\gamma$ but would otherwise induce the desired perception:
\begin{equation}
\gamma^* = PV^-(\Sigma^*)
\end{equation}

This trajectory corresponds to how legible motions~\cite{GenLegible} and explicable plans~\cite{exp} are considered.  
However, such an approach leads to cost-suboptimal behaviors.
In ASD, given the association between robot and its shadow during perception, the shadow can affect how the robot's behavior is perceived. 
Hence, we consider to merge the physical behavior from $\gamma$ with the accompanying information $\tau^*$ from $\gamma^* = \gamma^{*-} + \tau^*$ to generate a new  behavior $\gamma^D$, where $\tau_t^* = S(\gamma_t^{*})$, yielding: 
\begin{equation}
\gamma^D = \gamma^- + \tau^*
 \label{ASD}
\end{equation}

We hypothesize that such a new behavior, without changing the robot's physical trajectory $\gamma^-$, would move the perception of the robot's behavior towards $\Sigma^*$ as desired.
In other words, this method enables a robot to follow the cost-optimal trajectory $\gamma^-$ while creating the ``illusion'' that it is moving along a different trajectory $\gamma^*$.
It is clear that the maintenance of association between $\gamma^-$ and $\tau^*$ is crucial to the effectiveness of such illusions. 
Intuitively, when the robot and its shadow move in a non-coordinated way (e.g., moving in the opposite directions), the association may be broken and $\tau^*$'s influence on the perception would be lost.
In this paper, however, our main goal is to validate the effect of such illusions to manipulate perception. 
Hence, we provide a preliminary analysis of such an issue and will defer a detailed investigation to future work.

We also consider a baseline that uses explicit communication (referred to as BEC) to directly project $\gamma^{*-}$ (as $\epsilon$ in Fig. \ref{fig:illustrative}; see an example in Fig. \ref{desired_trajectory}) to influence the perception:

\begin{equation}
  \textcolor{black} {\gamma^{D\text{-}EC} = \gamma^- + \gamma^{*-}} 
    \label{DEC}
\end{equation}
It is worth noting a substantial difference between Eq. \eqref{ASD} and \eqref{DEC} lies in how the information is perceived. 
In Eq. \eqref{ASD}, the two parts are perceived integratively while they are perceived separately in Eq. \eqref{DEC} (see Fig. \ref{fig:illustrative}). 
We hypothesize that such a separation has a limited impact on perception. 
\begin{figure}
    \centering
    \caption{Flowchart of ASD}
    \label{fig:asd_model_flowchart}
\tikzstyle{startstop} = [rectangle, rounded corners, 
minimum width=3cm, 
minimum height=1cm,
text centered, 
draw=black, 
fill=purple!30,
text width=8cm,    
align=center]

\tikzstyle{io} = [trapezium, 
trapezium left angle=70, 
text width=8cm, 
trapezium right angle=100, 
minimum width=3cm, 
minimum height=1cm, text centered, 
draw=black, fill=blue!30]

\tikzstyle{process} = [rectangle, 
minimum width=3cm, 
minimum height=1cm, 
text centered, 
text width=6cm, 
draw=black, 
fill=orange!30]

\tikzstyle{decision} = [diamond, 
minimum width=3cm, 
minimum height=1cm, 
text centered, 
draw=black, 
fill=green!30]
\tikzstyle{arrow} = [thick,->,>=stealth]

\begin{tikzpicture}[node distance=1.6cm]

\node (in1) [startstop] {Input: $\gamma$ (optimal robot behavior) and $\Sigma^*$- \\ desired perception of the robot's behavior};
\node (pro1) [process, below of=in1] {Decompose robot behavior: $\gamma \leftarrow \gamma^- + \tau$. Where $\gamma^-$ is the physical behavior and $\tau$ is the natural shadow};
\node (pro2) [process, below of=pro1] {Compute another trajectory $\gamma^* = PV^-(\Sigma^*)$ that induces the desired perception};
\node (pro3) [process, below of=pro2] {Decompose desired perception behavior: $\gamma^* \leftarrow \gamma^{*-} + \tau^*$};
\node (pro4) [process, below of=pro3] {Merge the physical behavior from $\gamma$ with the shadow from $\gamma^*$ to generate $\gamma^D = \gamma^- +\tau^*$};

\node (out2) [startstop, below of=pro4] {Output:                                       
$\gamma^D = \gamma^- +\tau^*$  - New behavior generated by virtually projecting $\tau^*$};

\draw [arrow] (in1) -- (pro1);
\draw [arrow] (pro1) -- (pro2);
\draw [arrow] (pro2) -- (pro3);
\draw [arrow] (pro3) -- (pro4);
\draw [arrow] (pro4) -- (out2);
\end{tikzpicture}
\vskip-10pt
\end{figure}


\subsection{Case Studies}
\label{case}
Next, we present several applications of ASD that are later used to illustrate its effectiveness in manipulating perception. 

\begin{figure*}[t!]
\centering
\subfigure[ASD]{\includegraphics[scale=0.1]{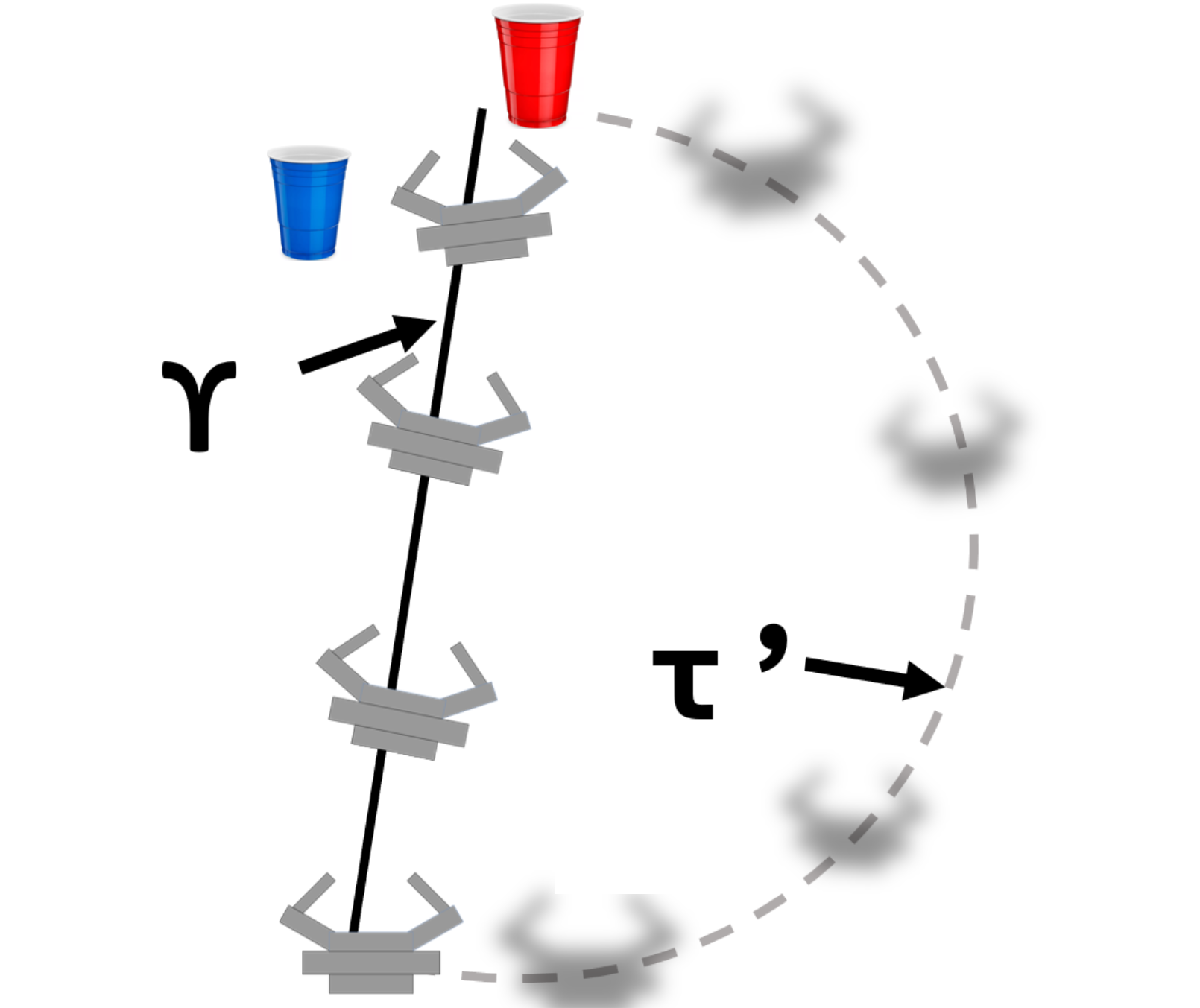}\label{legible_with_error}}
\subfigure[BEC]{\includegraphics[scale=0.1]{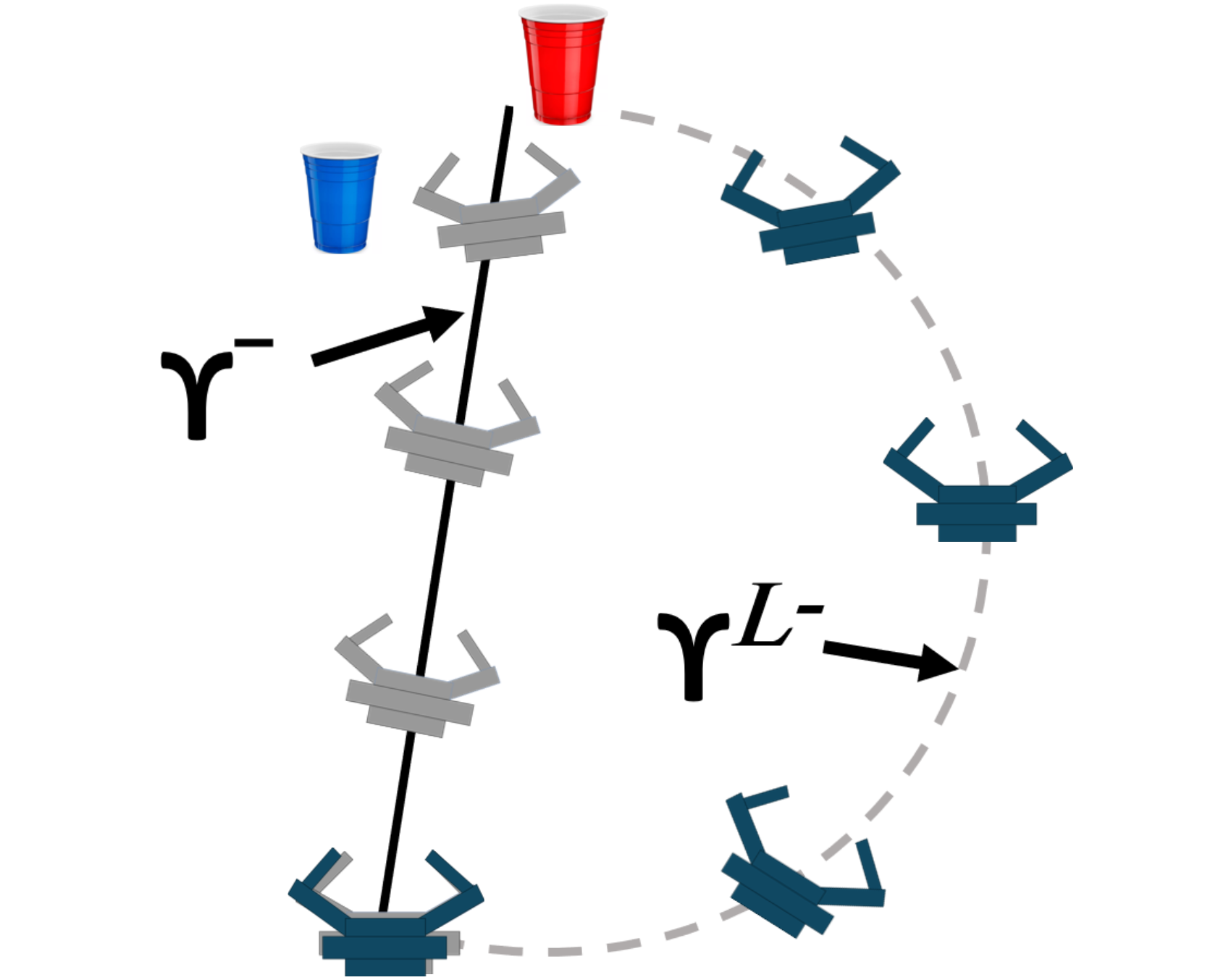}\label{Shad_with_error}}
\subfigure[BIC]{\includegraphics[scale=0.1]{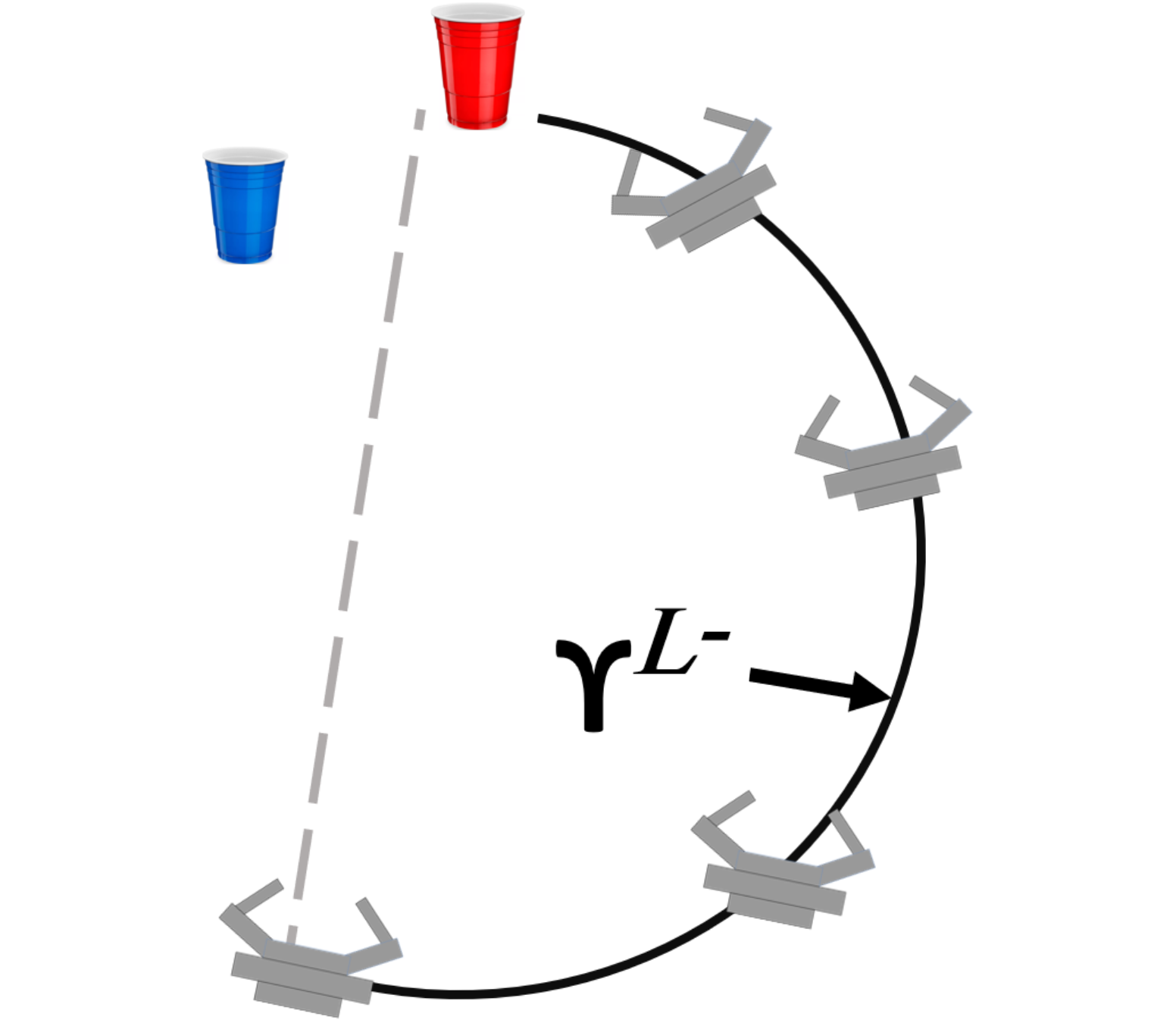}\label{hologram_with_error}}
\caption{Illustration of ASD, BEC (Eq. \eqref{DEC}), and BIC (as legible motion~\cite{legPred}), respectively, to achieve legible trajectory. 
}
\label{fig:errors}
\vskip-10pt
\end{figure*}

{\it Illusion of Motion:}
We begin with the simplest illusion of robot moving while it is not:
\begin{equation}
  \textcolor{black} {\gamma^D = \gamma^- + \tau^M} 
    \label{illusion_of_motion}
\end{equation}
where $\gamma^- = \bf{0}$ and $\tau^M$ is the shadow trajectory under the initial light orientation for a moving trajectory $\gamma^M$ ($\gamma_0^{M-} = \bf{0}$ and $\gamma_t^{M-} \not= {\bf{0}} \, (t > 0)$) , such that $\tau_t^M = S(\gamma_t^{M})$. 
In this case, the robot remains at its initial position throughout while \(\tau^M\) is added to create the illusion of motion. 



{\it Illusion of Legible Motion:} 
Given a cost-optimal trajectory $\gamma$ and a legible trajectory $\gamma^{L}$ based on  Eq.~\eqref{eq:Anca}, similarly, we compute  $\tau_t^L = S(\gamma_t^{L})$, and then combine to yield:

\begin{equation}
    \gamma^D = \gamma^- + \tau^L
    \label{legible_motion_illusion}
\end{equation}

We graphically illustrate ASD, legible motion (as a prior IVC method), and BEC in Fig.~\ref{fig:errors}.


{\it Illusion of Imminent Collision:} 
Armed with the lack of physical constraints,  
shadows can be used to create the illusion of imminent collision to facilitate timely identification and aversion of potential risks.
Such an effect is evident when considering the shadow of a person approaching you from a corner before the person becomes visible. 
To create the illusion of imminent collision, we generate $\gamma^{I}$ ahead of $\gamma$:
\begin{equation}
    \gamma_t^{I} = \gamma_{t + k}
    \label{future_state}
\end{equation}
where $k$ is an arbitrary time displacement. 
Then, we compute its shadow trajectory by $\tau_t^{I} = S(\gamma_t^{I})$, and add it to $\gamma^{-}$ as with other cases.
This enables ASD to essentially communicate the future state of a robot, such as introducing the perception that the robot  approaches a location before it actually does. 


\begin{figure*}
\centering
\subfigure[ASD]{\includegraphics[scale=0.083]{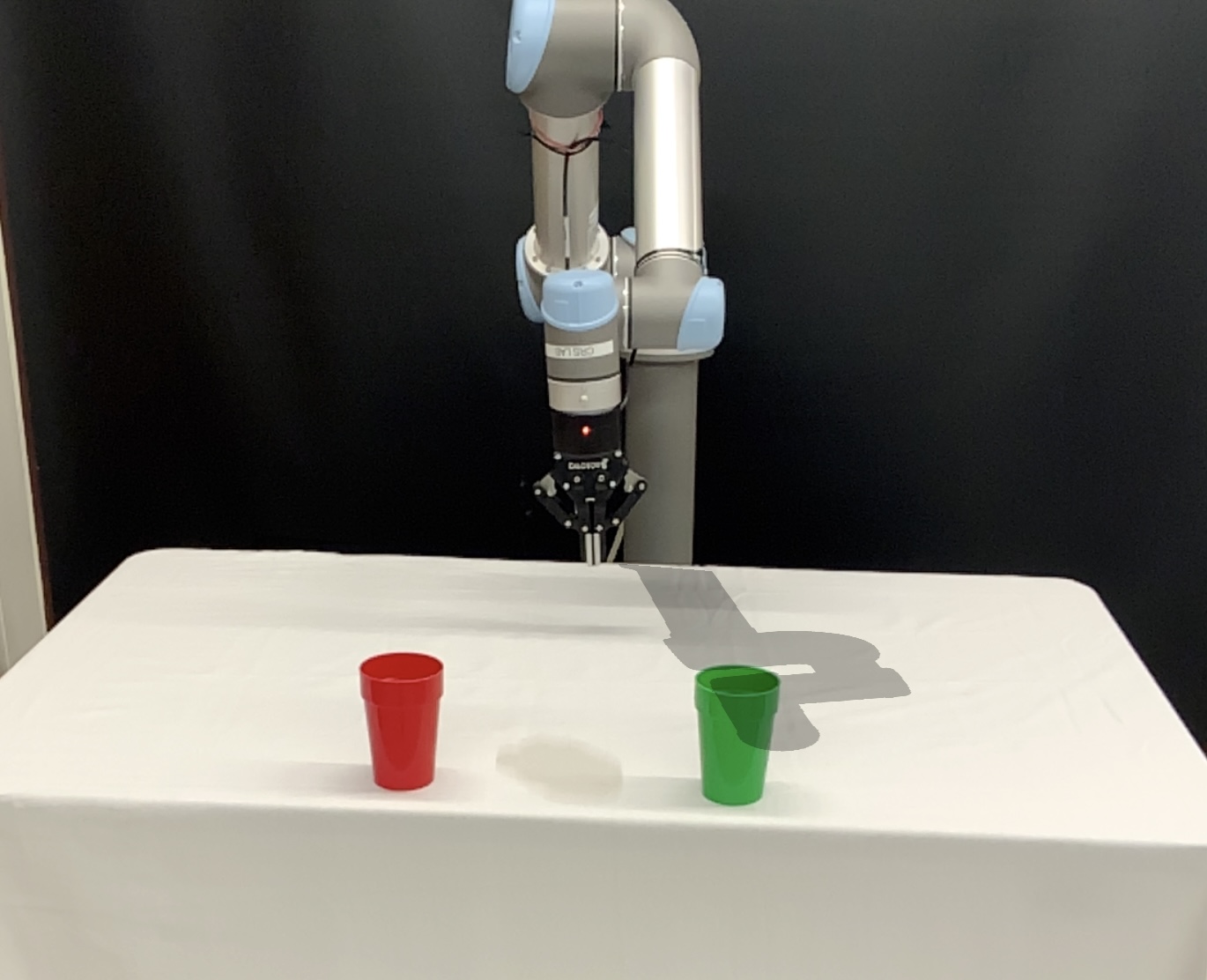}\label{shadow_legible_motion}}
\subfigure[BEC]{\includegraphics[scale=0.172]{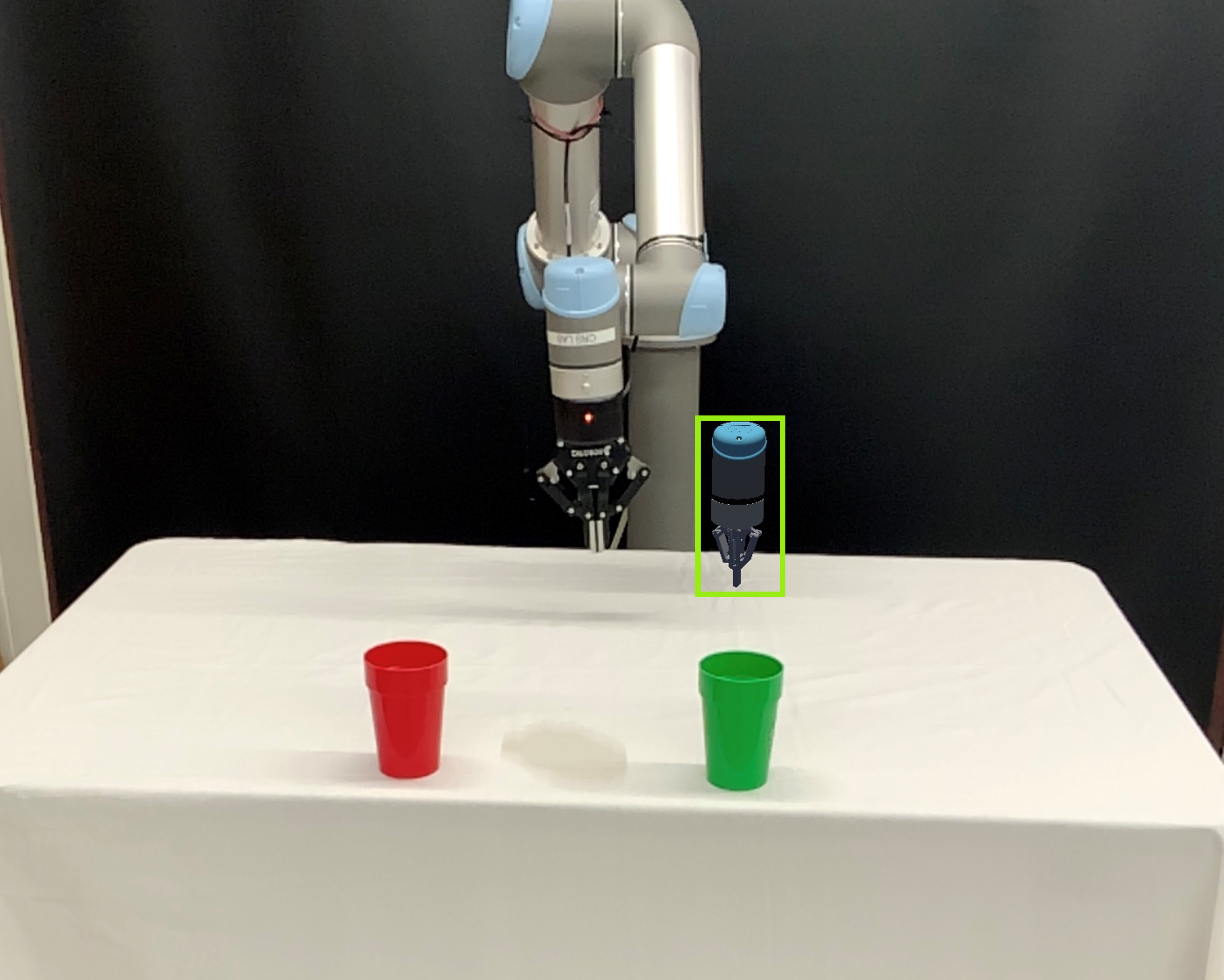}\label{hologram}}
\subfigure[BIC~\cite{legPred}]{\includegraphics[scale=0.083]{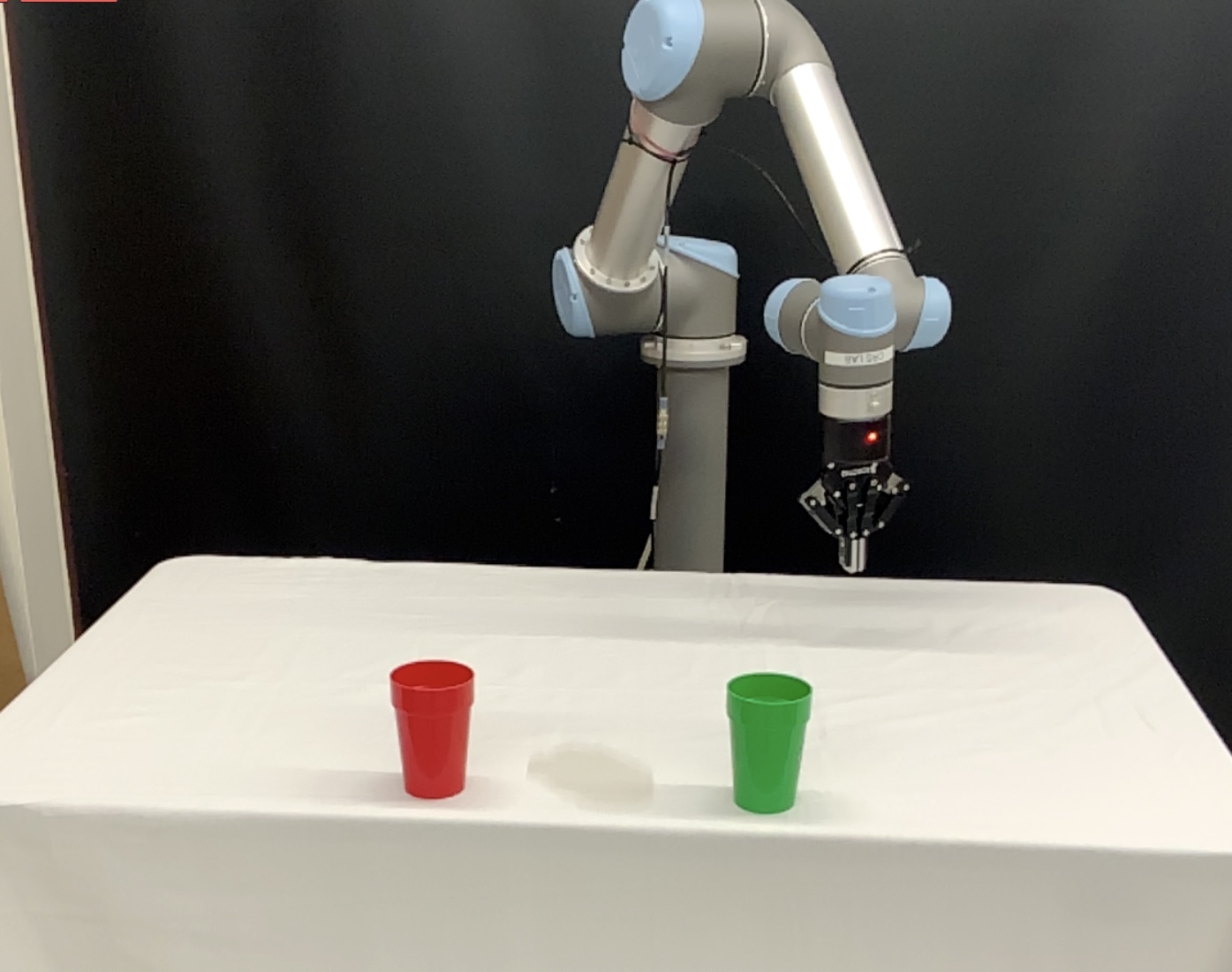}\label{ancalegible}}
\vskip-5pt
\caption{Experimental setup of the different methods for legible motion with the green cup as the goal (the front view).}
\label{fig:AllMethods}
\vskip-10pt
\end{figure*}

\section{Evaluation and Experimental Design}
\subsection{Hypotheses}
\noindent \textit{H1: ASD is effective at creating illusions that alter the perception of robot behavior.} \\
\noindent \textit{H2: ASD can maintain the (cost-)optimality of trajectories without sacrificing their understandability.} \\
\noindent \textit{H3: The association between the virtual shadow and  robot can be broken and in such cases the shadow's influence on the perception would be weaken or lost.}\\
\noindent \textit{H4: ASD is comparable with the best performing baselines in terms of mental workload.} \\
\noindent \textit{H5: ASD can be more informative than EC methods.} \\
\subsection{Experimental Design}
 
We compare ASD with BEC, and prior implicit visual communication (IVC) methods that require changing the robot behavior, referred to as BIC, under the three applications considered in Sec. \ref{case}.
Note that prior IVC methods are not always applicable for the applications considered, such as the illusion of motion and imminent collision, suggesting a limitation of theirs.
This is because changing the physical robot behavior to create those illusions makes them no longer illusions.
In such cases, we will simply use the original behavior as BIC.



\subsection{Experimental Setup}
Our experimental setup is a dining scenario that involves simple maneuvers of a robot manipulator (UR5), such as moving around on a dining table with cups or wine glasses on top. 
The (cost-)optimal physical behavior in all scenarios is the shortest trajectory from its initial position to the goal. 
To reduce the influence of the manipulator's posture and wrist rotation on perception, 
we restrict the wrist to translational motions except for the last evaluation (Sec. \ref{physical demonstration}). 

\subsubsection{Trajectory Control and Visualization}

 Once a trajectory is generated, it is provided either to the manipulator controller or the visual projection control system (for ASD and BEC)
 in the form of discretized waypoints for control and visualization. 

\subsubsection{Survey and Demographics}
We ran a series of between-subject studies for trajectories generated for the different methods and under the different applications discussed in Sec. \ref{case}. 
We created an AR app in Unity to record these trajectories and presented them in the survey. A total of 138 participants were initially recruited for the survey of which 64 were 21 years old or younger, 55 were between 22 and 30 years old and 19 were older than 30 years.
All participants completed the survey online under supervision in a room. 
21 responses were rejected because they were incomplete or failed the sanity checks. 
This resulted in coincidentally 
39 participants per method. 

\paragraph{Illusion of motion}
We created a scenario where we fixed the UR5 and moved only the virtual shadow according to $\tau^M$ in Eq. \eqref{illusion_of_motion} to create the illusion of motion for ASD. 
Participants were asked to indicate if the UR5 was moving and, if so, in which direction. 
The same scenario was also evaluated using BEC.

Furthermore, in the same scenario, we kept the UR5 stationary and gradually moved the virtual shadow some distance away such that it corresponded to a maximum of $45^\circ$ change to the angle of elevation \((\alpha\)). 
Participants were asked to indicate the moment they noticed that the robot had stopped moving. 
The same scenario was also evaluated using BEC since illusion of motion is only applicable to ASD and BEC. 
To minimize its impact on the evaluation of the two other illusions and subjective questions, the above activity for ASD and BEC participants was presented after the subjective questions were asked.

\paragraph{Illusion of legible motion}

We designed two scenarios (corresponding to a front and a side view to study the influence of different perspectives) where the UR5 was to move towards either a red or green cup (the goal) as shown in Fig. \ref{fig:AllMethods}.
The illusion of legible motion was created according to Eq. \eqref{legible_motion_illusion} where the virtual shadow followed the shadow of the legible trajectory while the UR5 followed the cost-optimal trajectory. Participants were asked to identify the goal of the robot as soon as they could. The same scenarios were also evaluated using BEC and BIC where BIC was implemented by the robot following the legible trajectory.


\begin{figure*}
\centering
\subfigure[ASD]{\includegraphics[width=0.45\columnwidth, height=0.36\columnwidth]{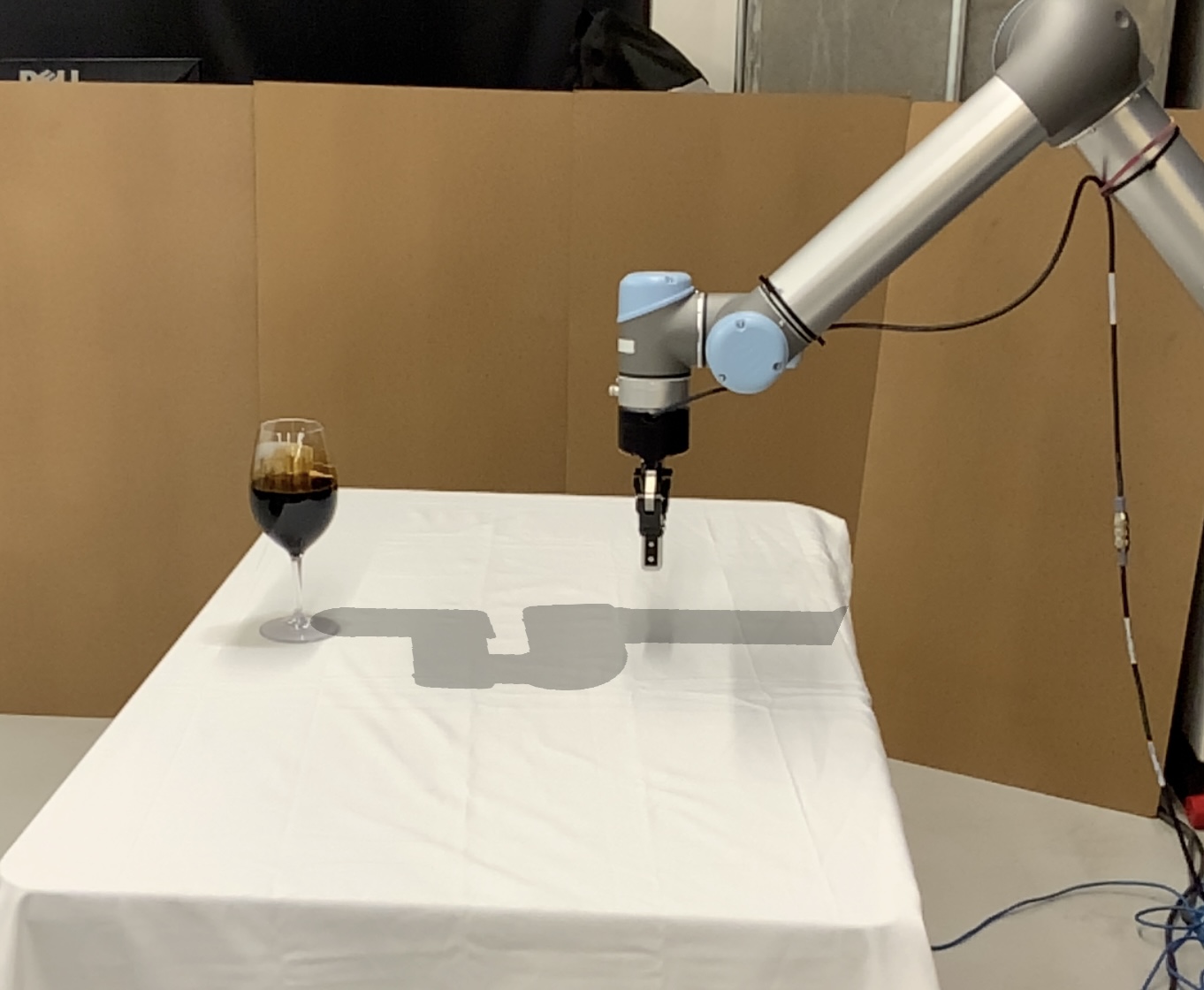}\label{Active_shadow_side}}
\subfigure[BEC]{\includegraphics[width=0.45\columnwidth, height=0.36\columnwidth]{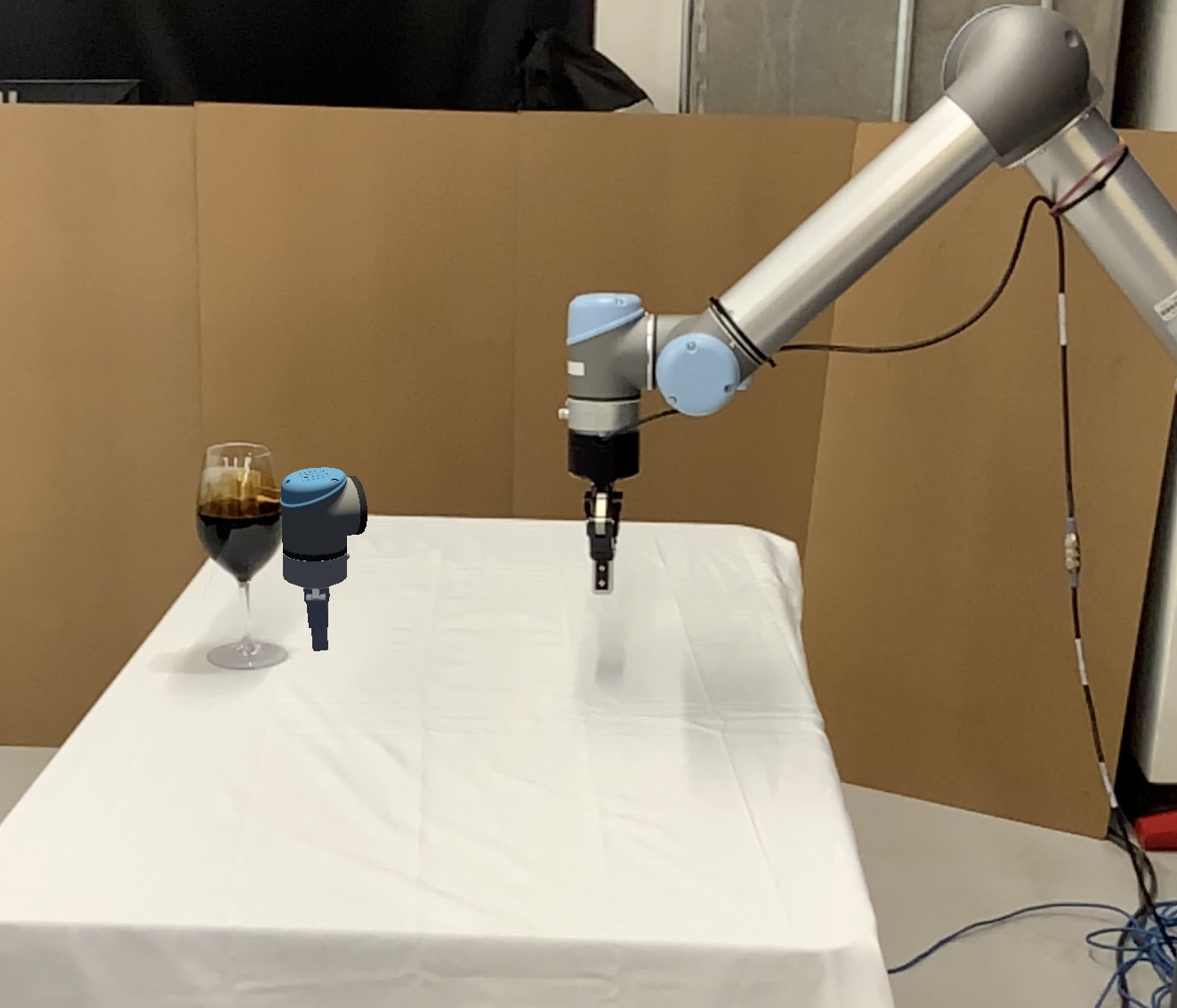}\label{Holo_side}}
\subfigure[BIC]{\includegraphics[width=0.45\columnwidth, height=0.36\columnwidth]{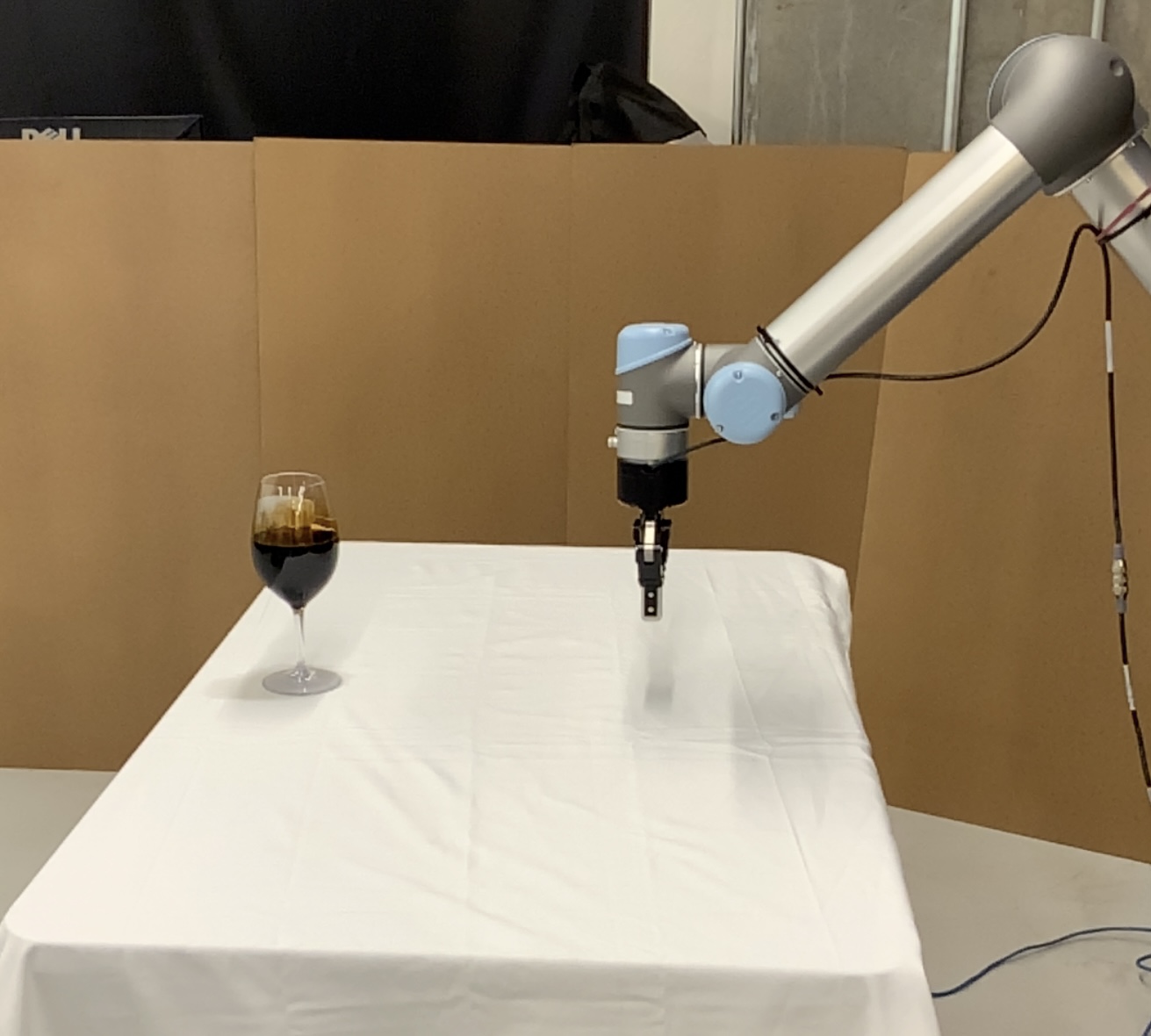}\label{Baseline_Side}}
\vskip-5pt
\caption{Illustration of the three different methods used in creating illusion of imminent collision (the side view).}
\label{fig:Side_view}
\vskip-15pt
\end{figure*}

\paragraph{Illusion of imminent collision}

We designed two scenarios (corresponding to a front and a side view) where the UR5 might appear to potentially knock over a glass of wine as it approached, as shown in Fig. \ref{fig:Side_view}. In these scenarios, the trajectories of ASD were computed based on Eq.~\eqref{future_state} where $\gamma$ simply moved towards the glass but stopped just short of reaching it.
Participants were asked to press a button to stop the robot whenever they felt it might knock over the glass. The same scenarios were also evaluated using BEC and BIC, where BIC is the robot's behavior without changes. 


To evaluate the different methods in terms of the induced mental workload, we conducted a NASA TLX subjective study at the end of the survey using a 7-point Likert scale (from 1 being the lowest to 7 being the highest) for each method. Participants were asked to rate how easy it was to determine the robot's intent (Q1), how intuitive and suitable the interaction was (Q2 and Q6), how demanding and difficult the tasks were (Q4 and Q5), and how confident they felt in their responses (Q3).

\section{RESULTS AND ANALYSES}

\begin{table*}[ht]
\caption{\textbf{Mean and Standard Deviation for Response Time (in seconds) for Legible Motion and Imminent Collision}}
\centering
\renewcommand{\arraystretch}{2}
\begin{tabular}
{p{0.09\linewidth}p{0.13\linewidth}p{0.17\linewidth}p{0.13\linewidth}p{0.13\linewidth}p{0.13\linewidth}p{0.13\linewidth}|p{0.07\linewidth}|}
\hline
\textbf{Method} &  Imminent Collision & Imminent Collision-Side & Legibility & Legibility-Side & $\zeta (cm^2)$ \\
\hline
\textbf{BIC} & 7.337 $\pm$ 3.08 &  6.181 $\pm$ 2.26 & 7.964 $\pm$ 2.88 &5.613 $\pm$ 2.03 &267.69  \\
\hline
\textbf{BEC} & 7.259 $\pm$ 3.02 & 7.140 $\pm$ 4.03 & 8.301$\pm$ 3.82 & 7.883 $\pm$ 3.27& 11.23  \\
\hline
\textbf{ASD} & 7.075 $\pm$ 3.08 &5.743 $\pm$ 2.51 & 9.056 $\pm$ 4.02 &5.715 $\pm$ 2.50 &11.78  \\
\hline
\end{tabular}
\label{tab:resultTable}
\vskip-10pt
\end{table*}

An alpha level of $0.05$ was used for all our statistical tests. 
\subsection{Results}
\subsubsection{Illusion of Motion} 

71.79\% of ASD participants believed the robot was moving in the direction of its shadow, even though the robot was stationary. In contrast, only 15.22\% of BEC participants believed the robot was moving in the direction of its hologram. 
This result validated H1 for illusion of motion.

As the distance between the virtual shadow and robot grew,
the association substantially weakened as shown in Fig.~\ref{fig:threshold}. 
At a distance that corresponded to a change to the angle of elevation ($\alpha$) of $15^{\circ}$, 
60\% of the respondents perceived that robot was moving along with the virtual shadow while only 10\% and 7.5\% indicated the same when $\alpha$ increased to $30^{\circ}$ and $45^{\circ}$, respectively.
This result validates H3. 
Conversely, with BEC, only about 15\% of the participants perceived the robot as moving along the hologram, regardless of the distance between the robot and the hologram converted to an equivalent $\alpha$, respectively. 
This result further validates H1 compared to BEC and suggested that the strength of association in ASD was much stronger than BEC as expected.
In this regard, the seemingly comparable performance of BEC in creating legible motion next may not be attributed to the association between the robot and the hologram but rather that the hologram was seen as a goal signal. 
We will study such an effect in more depth and propose methods to maintain the association in future work. 

\begin{figure}
\centering
\begin{tikzpicture}
\begin{axis}[
    ybar,
    xtick=data, 
    symbolic x coords={$\epsilon = 15$, $\epsilon = 30$, $\epsilon = 45$},
    legend pos=north east,
    legend columns=-1,
    legend style={font=\footnotesize},
    width=0.4\textwidth,
    height=0.24\textwidth,
    bar width=7, 
    extra y tick style={grid=major, grid style={dotted, black}}, 
    x tick label style={font=\small},
]
\addplot[fill = gray, error bars/.cd, y dir=both, y explicit] coordinates {
   ($\epsilon = 15$,0.6)
   ($\epsilon = 30$,0.1)
   ($\epsilon = 45$,0.075)
   };
\addplot+[fill = orange, error bars/.cd, y dir=both, y explicit] coordinates {
    ($\epsilon = 15$,0.152) 
   ($\epsilon = 30$,0.152)
   ($\epsilon = 45$,0.152)
};
\legend{ ASD, BEC}
\end{axis}
\end{tikzpicture}
\caption{Percentage of participants that indicated an observed motion in the robot at different $\epsilon = \Delta(\alpha)$.}
\label{fig:threshold}
\vskip-13pt
\end{figure}

\subsubsection{Illusion of Legible Motion} 

As shown in Table \ref{tab:resultTable}, using the front view, BIC was the most legible of all three methods,  likely due to its self-explanatory physical behavior that however deviates from the cost-optimal trajectory. 
Pair-wise \textcolor{black}{one-way} 
ANOVA tests for legibility between ASD and BIC, ASD and BEC, and BEC and BIC yielded scores of (\textit{F = 4.3, F-crit = 1.84, p = 0.009}), (\textit{F = 4.04, F-crit = 1.47, p = 0.021}) and (\textit{F = 4.05, F-crit = 1.79, p = 0.013}), respectively. 
ASD did not perform as well here.  
This result seemed to suggest that the viewing angle played a role in using ASD to introduce legible motion, especially in combination with the result below, which we will analyze in future work.

For the side view, ASD outperformed BEC.
Pair-wise one-way ANOVA tests between ASD and BIC, ASD and BEC, and BEC and BIC yielded scores of (\textit{F = 8.18, F-crit = 3.98, p = 0.047}), (\textit{F = 3.96, F-crit = 3.03, p = 0.030}), and (\textit{F = 7.86, F-crit = 4.06, p = 0.017}), respectively.
Note that our goal here was not to show that ASD could outperform BEC or BIC in terms of legibility, but rather that it could achieve a comparable effect of facilitating behavior understandability
when compared side-by-side with these competing methods, one being the state-of-the-art and specifically designed for generating legible motions.

We also compared the cost of trajectories for each method. The cost, calculated as the sum of squared distance errors between the optimal and executed trajectory, denoted by ($\zeta$), indicates that while BIC scored the highest in legibility, it sacrificed optimality substantially. In contrast, ASD maintained a comparable level of legibility without compromising optimality. These results in combination validate H2.

\begin{figure}[t] 
\centering
\subfigure[Robot goal informed choice]{
    \includegraphics[width=0.45\columnwidth, height=0.35\columnwidth]{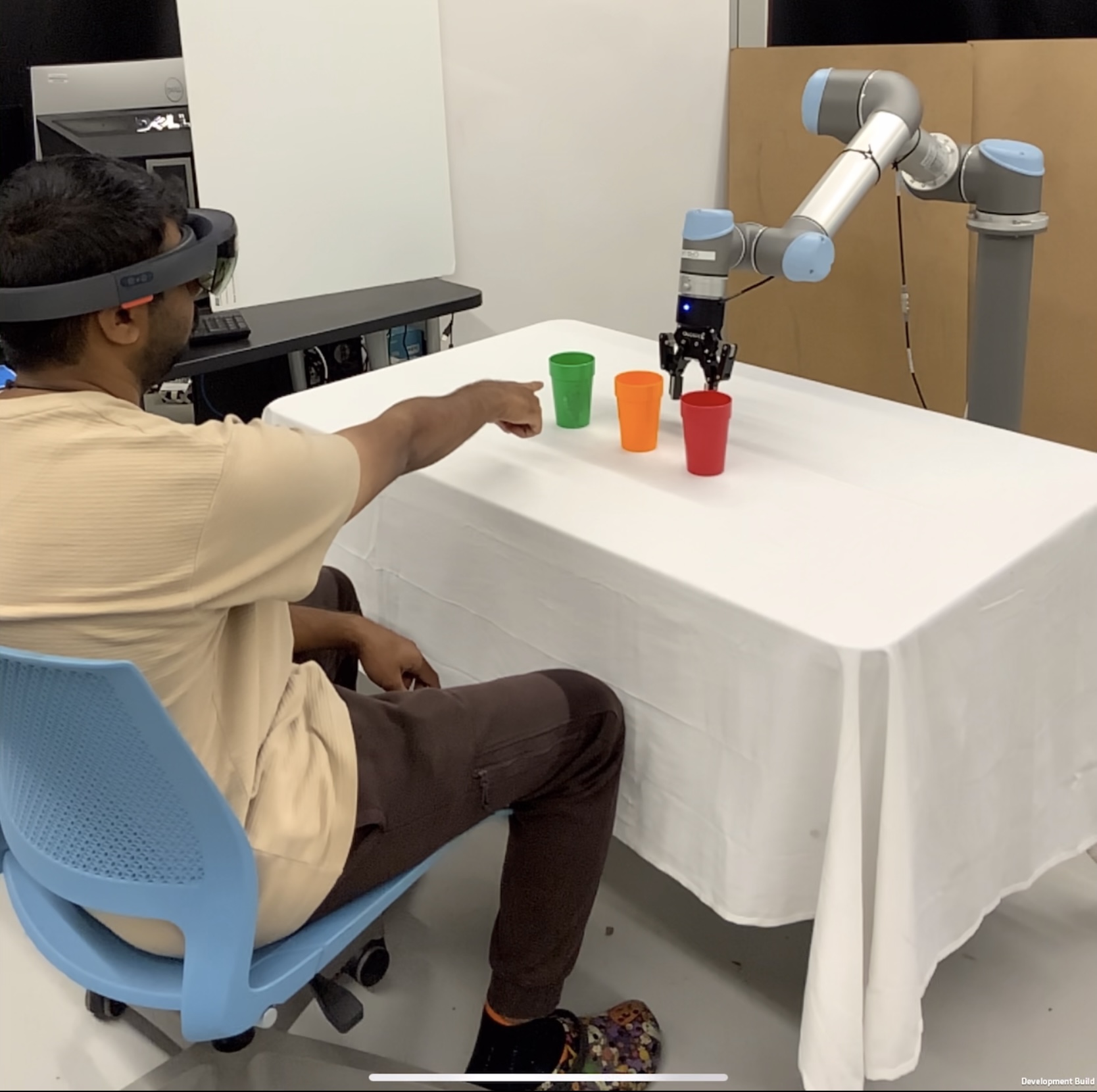}
    \label{fig:collision_b}
}
\hfill
\subfigure[Unsafe collaboration]{
    \includegraphics[width=0.45\columnwidth, height=0.35\columnwidth]{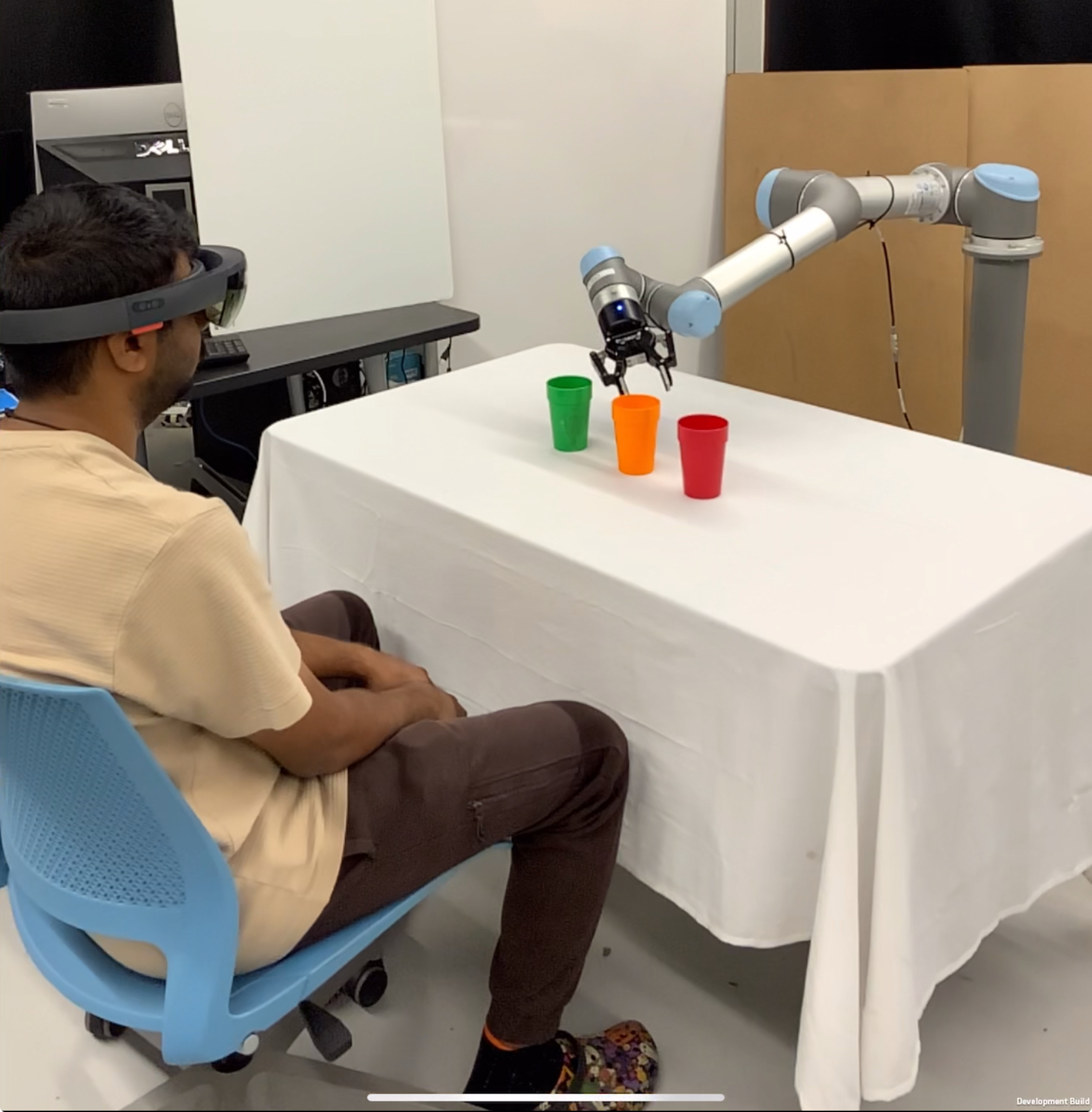}
    \label{fig:collision_up}
}
\vskip-5pt
\caption{\textcolor{black}{(a) The robot moved and stopped in front of the orange cup in the middle. 
The human chose to pick the green cup since there was no further information to gauge safety. 
However, due to the gripper orientation during pickup (shown in b), choosing the green would be an unsafe choice: the butt end of the gripper would be too close to the green cup.}}
\label{fig:coll}
\vskip-5pt
\end{figure}

\begin{figure}[t] 
\centering
\subfigure[ASD informed choice]{
    \includegraphics[width=0.45\columnwidth, height=0.35\columnwidth]{Pictures/shad_safety_b_.png}
    \label{fig:Shad_safety_b}
}
\hfill
\subfigure[Safe collaboration]{
    \includegraphics[width=0.45\columnwidth, height=0.35\columnwidth]{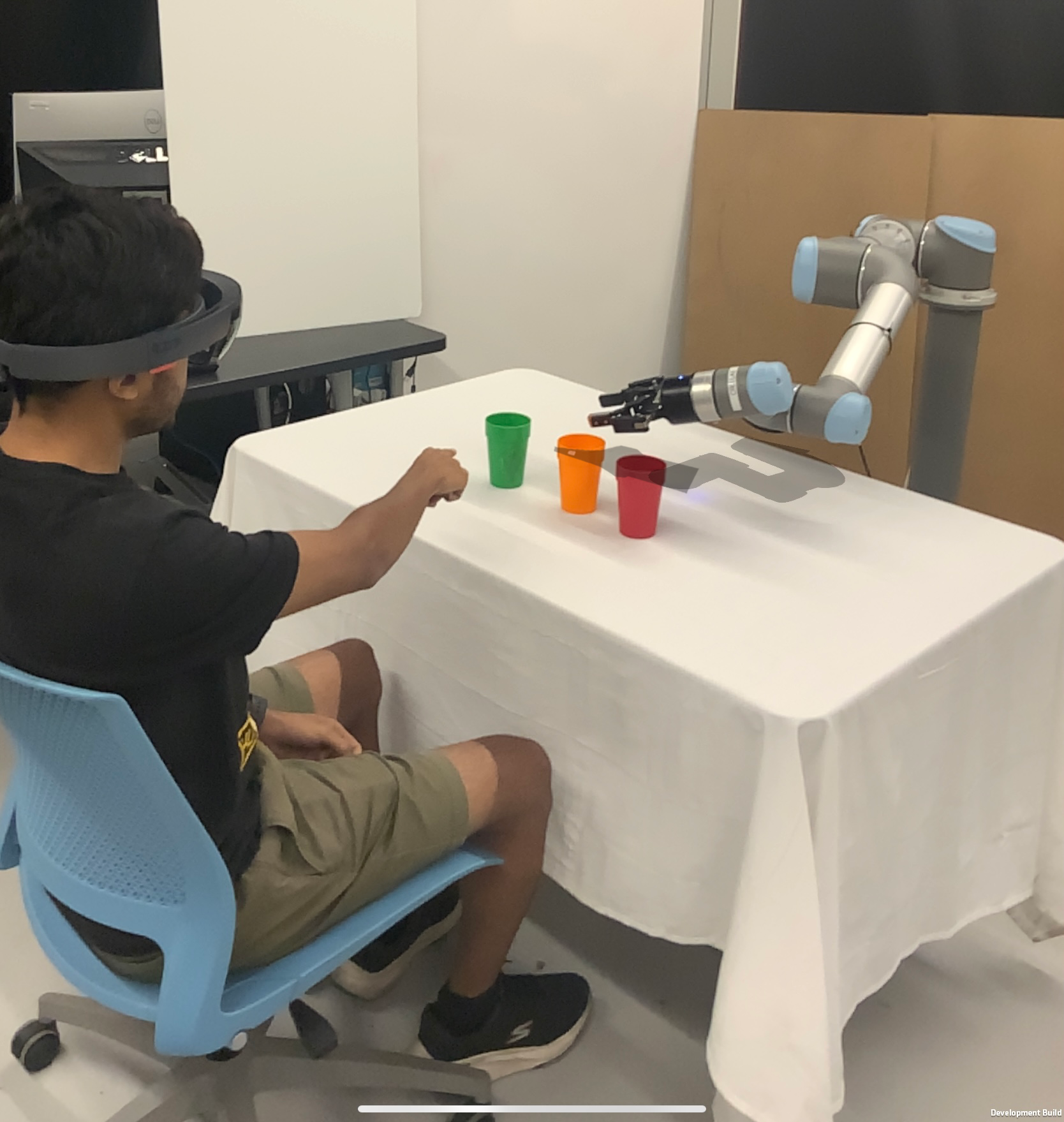}
    \label{fig:Shad_safety_up}
}
\vskip-5pt
\caption{\textcolor{black}{(a) - The human after observing the movement of the shadow realized that the gripper would be orientated such that the green cup would be the safer choice and confidently chose to pick up the green cup.
(b) The gripper's orientation confirmed that the green cup was truly the safer choice.}}
\label{fig:confid}
\vskip-5pt
\end{figure}

\subsubsection{Illusion of Imminent Collision}

Results in Table~\ref{tab:resultTable} show that ASD facilitated faster detection of potential collisions than BEC or BIC. 
Pair-wise one-way ANOVA tests 
between ASD and BIC, ASD and BEC, and BEC and BIC yielded scores of (\textit{F = 3.96, p = 0.02, F-crit = 3.29}), (\textit{F = 3.98, p = 0.040, F-crit = 3.53}), and (\textit{F = 4.09, p = 0.049, F-crit = 3.33}) for the front view,  
and 
(\textit{F = 3.97, p = 0.037, F-crit = 3.14}), (\textit{F = 3.97, p = 0.024, F-crit = 3.53}), and (\textit{F = 4.09, p = 0.027, F-crit = 3.33}) for the side view.


It is interesting to see that although the robot's behavior remained unchanged for BIC, participants detected potential collisions faster than those in BEC in the side view.
 This is likely due to the lack of association between the hologram and the robot. Consequently, the hologram in the side view provided a point of reference for better estimating the distance between the robot and the hologram/collision, 
 resulting in undesirable late responses. 
The result here validates H1 for creating illusions of imminent collision. 

\subsection{Mental Workload}
The NASA-TLX result is presented in Fig.~\ref{fig:subjectiveResults}. 
We conducted one-way ANOVA tests for each of the questions asked (\textit{F = 3.43, p = 0.03, F-crit = 3.06}), (\textit{F = 3.47, p = 0.02, F-crit = 3.06}), (\textit{F = 3.38, p = 0.04, F-crit = 3.06}), (\textit{F = 3.10, p = 0.036, F-crit = 3.06}), (\textit{F = 3.46, p = 0.049, F-crit = 3.06}),  (\textit{F = 3.43, p = 0.034, F-crit = 3.06}) for Intent, Intuitive, Confidence, Demanding, Difficulty and Suitable respectively. 
Overall, the results indicated that ASD was generally comparable with the baselines and exceeded them in some category, in particular Demand, thus validating H4. 

\begin{figure}
\centering
\begin{tikzpicture}
    \begin{axis}[
        ybar,
        xtick=data,
        symbolic x coords={Intent, Intuitive, Confidence,Demand, Difficult, Suitable}, 
        legend pos=north east,
        legend columns=-1,
        legend style={font=\footnotesize},
        width=0.54\textwidth,
        height=0.26\textwidth,
        bar width=6.5, 
        extra y ticks={1, 2, 3, 4, 5}, 
        extra y tick style={grid=major, grid style={dotted, black}}, 
        x tick label style={font=\small},
        nodes near coords align={vertical},
    ]

    \addplot+[fill = blue, error bars/.cd, y dir=both, y explicit] coordinates {
        (Intent,5.589) +- (0.9,0.9)
        (Intuitive,5.410) +- (1.1,1.1)
        (Confidence,5.692) +- (1.0,1.0)
        (Demand,4.231) +- (0.5,0.5)
        (Difficult,4) +- (0.3,0.3)
        (Suitable,4.615) +- (1.4,1.4)
    };

    \addplot[fill = orange, error bars/.cd, y dir=both, y explicit] coordinates {
        (Intent,4.742) +- (1.5,1.5)
        (Intuitive,4.774) +- (1.2,1.2)
        (Confidence,5.0) +- (1.4,1.4)
        (Demand,4.226) +- (1.3,1.3)
        (Difficult,4.279) +- (1.3,1.3)
        (Suitable,3.871) +- (1.8,1.8)
    };

    \addplot[fill = gray, error bars/.cd, y dir=both, y explicit] coordinates {
        (Intent,5.406) +- (1.1,1.1)
        (Intuitive,5.089) +- (1.2,1.2)
        (Confidence,5.582) +- (1.1,1.1)
        (Demand,4.551) +- (1.5,1.5)
        (Difficult,4.335) +- (1.6,1.6)
        (Suitable,4.753) +- (1.5,1.5)
    };
    
    \legend{BIC, BEC, ASD}
    \draw (axis cs:Demand,5.2) node {*};
    \draw (axis cs:Intent,5.4) node {*};
    \draw (axis cs:Intuitive,5.4) node {*};
    \draw (axis cs:Suitable,4.4) node {*};
    \draw (axis cs:Difficult,4.6) node {***};

    \end{axis}

\end{tikzpicture}
\caption{Comparison of mental workload. ``Demand'' and ``Difficult'' were inverted to be consistent with others such that higher values are better. Significance level is also shown for the ANOVA tests: $*$ for $0.05$ and $***$ for $0.001$. 
}
\label{fig:subjectiveResults}
\vskip-15pt
\end{figure}

\subsection{Informativeness of ASD} \label{physical demonstration}
To demonstrate that ASD can be more informative for robot behavior, we created a scenario similar to the motivating scenario (see Figs. \ref{fig:motivating}, \ref{fig:coll} and \ref{fig:confid})  
where a human and a robot were tasked with picking evenly spaced cups from a shared workspace on a table. For safety, the participant was instructed to always select the cup that was the farthest from the robot manipulator. Participants were informed that the robot would be always picking the cup it paused in front of.
In all trials, the robot moved in a straight line from its home position and paused directly in front of the middle cup before picking.  Participants were then asked to choose one of the two remaining cups before the robot resumed picking the middle cup. To quantify safety, we assigned 1 safety point if the human chose the cup closest to the robot, and 3 safety points for choosing the cup that is farthest from the robot.
We tested two methods:
\begin{itemize}
    \item BEC (no change to behavior). However, the participants were verbally informed (EC) of the robot's choice before the robot began executing its trajectory.
    \item ASD: robot shadow projected several seconds ahead of its actual motion as in illusion of imminent collision.
\end{itemize}
A total of 23 participants (11 in BEC and 12 in ASD) took part in the evaluation. 
All wore Microsoft HoloLens to view the workspace.
We conducted t-test to compare safety scores between the BEC and ASD conditions. Participants in the ASD condition (M = 2.67, SD = 0.65) scored significantly higher than those in the BEC condition (M = 1.73, SD = 0.79), t(21) = 2.51, p = 0.020, indicating that ASD led to improved safety outcomes.
(Notably, one BEC participant who made the safe choice later commented that it was selected randomly when asked about their reasoning.) 
This result thus validates H5. 
Although this evaluation only studied a simple scenario, 
we believe that it is representative of many similar situations that can occur frequently with proximal human-robot collaboration. 

We attribute this result to the fact that while the robot’s goal was revealed in BEC, goal alone was insufficiently informative about the robot's behavior, such as whether the robot would choose to pitch left or right for the pickup in this evaluation.  
On the other hand, ASD provided more information due to constant shadow projection. 
Such projection foreshadowed the robot's future state throughout the execution for the participants to better anticipate potential risks.  
ASD however relies on its IVC nature to be practical.

\section{Conclusion}
This paper introduced a novel type of implicit visual communication (IVC) with unique characterization, referred to as active shadowing (ASD).
It bridged a gap left by the prior IVC and explicit communication methods.
ASD manipulated a behavior's naturally accompanying information (shadow) to influence how it would be perceived,
resulting in a discrepancy between the physical and its perceived behavior.
Building on a prior method for generating realistic virtual shadows, we implemented (ASD) via augmented reality. 
We explored several applications of ASD where it could generate different illusions. 
Our extensive human subject studies validated that ASD effectively created illusions that enhanced understandability without compromising optimality or mental workload. We also demonstrated that ASD could be more informative than EC methods due to its constant information projection, and identified and analyzed a key limitation of ASD to address in future work.



\bibliographystyle{unsrt}
\bibliography{main}

\end{document}